\documentclass[runningheads]{llncs}

\usepackage{graphicx}
\usepackage{algorithm}
\usepackage{algorithmic}
\usepackage{booktabs}
\usepackage{amsmath}
\usepackage{multirow}
\usepackage{multicol}
\usepackage{bm}
\usepackage[title]{appendix}
\usepackage{xcolor}
\usepackage{subcaption}
\usepackage{comment}
\usepackage[colorlinks,
citecolor=blue,
linkcolor=blue,
urlcolor=blue
]{hyperref}
\begin{document}
	%
	
	\title{LRTD: Long-Range Temporal Dependency based Active Learning for Surgical Workflow Recognition}
	
	\titlerunning{LRTD based Active Learning for Surgical Workflow Recognition}
	
	\author{Xueying Shi$^{\dagger}$ \and Yueming Jin$^{\dagger}$ \and Qi Dou \and Pheng-Ann Heng}
	\institute{
		Department of Computer Science and Engineering \\
		The Chinese University of Hong Kong, Hong Kong, China\\
		Email: xyshi,ymjin,qdou,pheng@cse.cuhk.edu.hk\\}
	%
	\maketitle          
\let\thefootnote\relax\footnote{$^{\dagger}$ joint first authors}
\begin{abstract}
\textbf{}\\
$Purpose $
Automatic surgical workflow recognition in video is an essentially fundamental yet challenging problem for developing computer-assisted and robotic-assisted surgery.
Existing approaches with deep learning have achieved remarkable performance on analysis of surgical videos, however, heavily relying on large-scale labelled datasets.
Unfortunately, the annotation is not often available in abundance, because it requires the domain knowledge of surgeons. Even for experts, it is very tedious and time-consuming to do a sufficient amount of annotations.\\
$Methods$
In this paper, we propose a novel active learning method for cost-effective surgical video analysis.
Specifically, we propose a non-local recurrent convolutional network (NL-RCNet), which introduces non-local block to capture the long-range temporal dependency (LRTD) among continuous frames.
We then formulate an intra-clip dependency score to represent the overall dependency within this clip.
By ranking scores among clips in unlabelled data pool, we select the clips with weak dependencies to annotate, which indicates the most informative ones to better benefit network training. \\
$Results$                                        
We validate our approach on a large surgical video dataset (Cholec80) by performing surgical workflow recognition task.
By using our LRTD based selection strategy, we can outperform other state-of-the-art active learning methods who only consider neighbor-frame information.
Using only up to 50\% of samples, our approach can exceed the performance of full-data training.\\
$Conclusion$
By modeling the intra-clip dependency, our LRTD based strategy shows stronger capability to select informative video clips for annotation compared with other active learning methods, through the evaluation on a popular public surgical dataset. The results also show the promising potential of our framework for reducing annotation workload in the clinical practice.

\end{abstract}
\section{Introduction}
Computer-assisted surgery and robotic-assisted surgery have been dramatically developed in recent years towards powerful support for the demanding scenarios of modern operating theatre, which is with highly complicated and extensive information for the surgeon~\cite{cleary2005or,james2007eye}.
Automatic surgical workflow recognition is a fundamental and crucial visual perception problem for computer-assisted surgery, which can enhance cognitive understanding of the surgical procedures in operating rooms~\cite{bricon2007context,dergachyova2016automatic}.
With accurate recognition of the surgical phases from endoscopy videos of minimally invasive surgery, a wide variety of downstream applications can be benefited from such context-awareness.
For instance, intra-operative recognition helps generate adequate notifications and alter future complications, by detecting rare cases and unexpected variations~\cite{bouget2015detecting,dergachyova2016automatic}.
Real-time phase identification can potentially support the decision making, the arrangement of team collaboration and the surgical process optimization during intervention~\cite{quellec2014real,forestier2015automatic,bouget2017vision}.
It can also assist to automatically index video database for surgical report documentation, which contributes developing post-operative tools for purposes of archiving, skill assessment and surgeon training~\cite{ahmidi2017dataset1,zappella2013surgical}.
In this regard, enhancing automatic workflow recognition of surgical procedure is essential in computer-assisted surgery for improving surgeon performance and patient safety.

The convolutional neural network (CNN) and recurrent neural network (RNN) have been widely utilized for workflow recognition from surgical video, as well as demonstrated their appealing efficacy of modeling spatio-temporal features for this task.
Existing successes achieved by deep learning models for workflow recognition are mostly based on fully supervised learning using frame-wise annotations~\cite{twinanda2016endonet,jin2017sv,jin2019multi}.
For instance, Twinanda et al.~\cite{twinanda2016endonet}
build a CNN to capture visual information of each frame, followed by a hierarchical hidden markov model (HMM) for temporal information refinement.
Jin et al.~\cite{jin2017sv} design an end-to-end recurrent convolutional model to jointly extract spatio-temporal features of video clips, where
a CNN module is used to capture frame-wise visual information, and a LSTM (i.e., long short term memory) module is utilized for the clip-wise sequential dynamics modeling.
However, these methods heavily relied on a large amount of data with extensive annotations to train the network.
Notably, the frame-wise annotations for surgical videos are 
quite expensive, as it requires expert knowledge and is highly time-consuming and tedious, especially when surgery duration lasts for hours.

With increasing awareness of the impediment from unavailability of large-scale labeled video data,
some works investigate semi-supervised learning to reduce annotation cost~\cite{yengera2018less,bodenstedt2017unsupervised,ross2018exploiting,funke2018temporal,yu2018learning}.
It can assist network training and promote prediction performance, with the demonstration that networks can learn a representation of certain inherent characteristics of the data, by first being trained towards the generated labels with the auxiliary task~\cite{doersch2017multi}.
Other semi-supervised methods use self-supervision with only a small portion of available labels~\cite{yengera2018less}.
Unfortunately, such semi-supervised methods could not make full use of the annotation workload, because data to be labelled are not carefully selected.
In addition, the current performance of semi-supervised learning is still less competitive to the fully supervised learning, which impedes clinical application in practice.
%

Instead, we explore sample mining techniques to incrementally enlarge the annotated database, so as to achieve state-of-the-art workflow recognition accuracy with minimal annotation cost.
We investigate the direction of active learning
\cite{settles.tr09}, which has been frequently revisited in the deep learning era to learn models in a more cost-effective way.
Its effectiveness has been verified by the successes of some medical image analysis scenarios (e.g., myocardium segmentation from MR image, grand segmentation from pathological data and disease classification from chest X-ray~\cite{mahapatra2018efficient,aaai2019ra,yang2017suggestive,zhou2017fine}), while less studied in the context of surgical video analysis.
The current state-of-the-art work Bodenstedt et al.~\cite{bodenstedt2019active} uses active learning to iteratively select a bunch of representative surgical sequences to annotate and progressively promote the workflow recognition performance.
They first estimate the uncertainty of each frame according to the likelihoods predicted by a recurrent deep Bayesian network (DBN).
The method then divides each video into segments with a length of five minutes, and select the most uncertain segments by averaging or maximizing the predictive entropy of all the frames within a segment.
High uncertainty verifies that the segments are hard and challenging for the network to recognize, while on the other hand, demonstrating their highly informative characteristic. 
Bodenstedt et al. assume that these samples are the most informative ones for annotation query, as they are key to learn the model more effectively and efficiently.

However, this previous active learning strategy selects video clips according to frame-wise uncertainty, where the uncertainty is first calculated separately for each single frame and then do the straightforward average and maximum operation to represent the entire clip.
Given that the surgical video is actually a form of sequential data, leveraging the cross-frame dependency to calculate the intra-clip dependency for sample selection are crucial for accurate workflow recognition.
Modeling the frame dependency within video clips can help to better identify the severe blur and noise samples which normally show the weak dependency with common surgical scenes.
It can also help to select the clips with significant intra-class variance, whose dependency are quite low.
Moreover, if there exist strong dependency within one clip, there is no need for network to be trained with the entire clip as there exist massive abundant information in such clip.
We incorporate the non-local operations which can capture long-range temporal dependency towards time steps~\cite{wang2018non}. Recurrent operations like LSTM process a local neighborhood in time dimension, thus long-range dependencies can only be captured when these operations are applied repeatedly, propagating signals progressively through the data.
However, repeating local operations has several limitations.
It is computationally inefficient and causes optimization difficulties.
These challenges further make multi-hop dependency modeling difficult, e.g., when information need to be delivered back and forward between distant time steps.

In this paper, we propose a novel active learning method to improve annotation efficiency for workflow recognition from surgical videos.
We design a non-local recurrent convolutional network (NL-RCNet), which builds the non-local operation block on top of a CNN-LSTM framework to capture long-range temporal dependency (LRTD) within video clips.
Such long-range temporal dependency can indicate the cross-frame dependencies among all the frames in a clip, without the limitation of time intervals.
Based on the constructed dependency matrix of a clip, we propose to calculate a intra-clip dependency score to represent the overall dependency of this clip.
By ranking scores of available video clips in the unlabelled data pool, we select the clips with lower scores and weaker dependencies to annotate, which are more informative to better benefit the network training.
To the best of our knowledge, we are the first to model clip-wise dependency for sample selection in active learning related to surgical video recognition tasks. Opposed to other approaches, which select the complete videos or individual frames, we aim to select the clips of 10 consecutive frames sampled at 1 fps.
We extensively validate our proposed NL-RCNet on a popular public surgical video dataset of Cholec80.
Our approach achieves superior performance of workflow recognition over existing state-of-the-art active learning methods.
By only requiring labeling 50\% clips, our method can surpass fully-supervised counterpart, which endorses the potential value in clinical practice.
Code for our proposed approach will be publicly available at \url{https://github.com/xmichelleshihx/AL-LRTD}.
\section{Method}
\begin{figure}[t!]
	\centering
	\label{figure1b}
	{\includegraphics[width=0.8\linewidth]{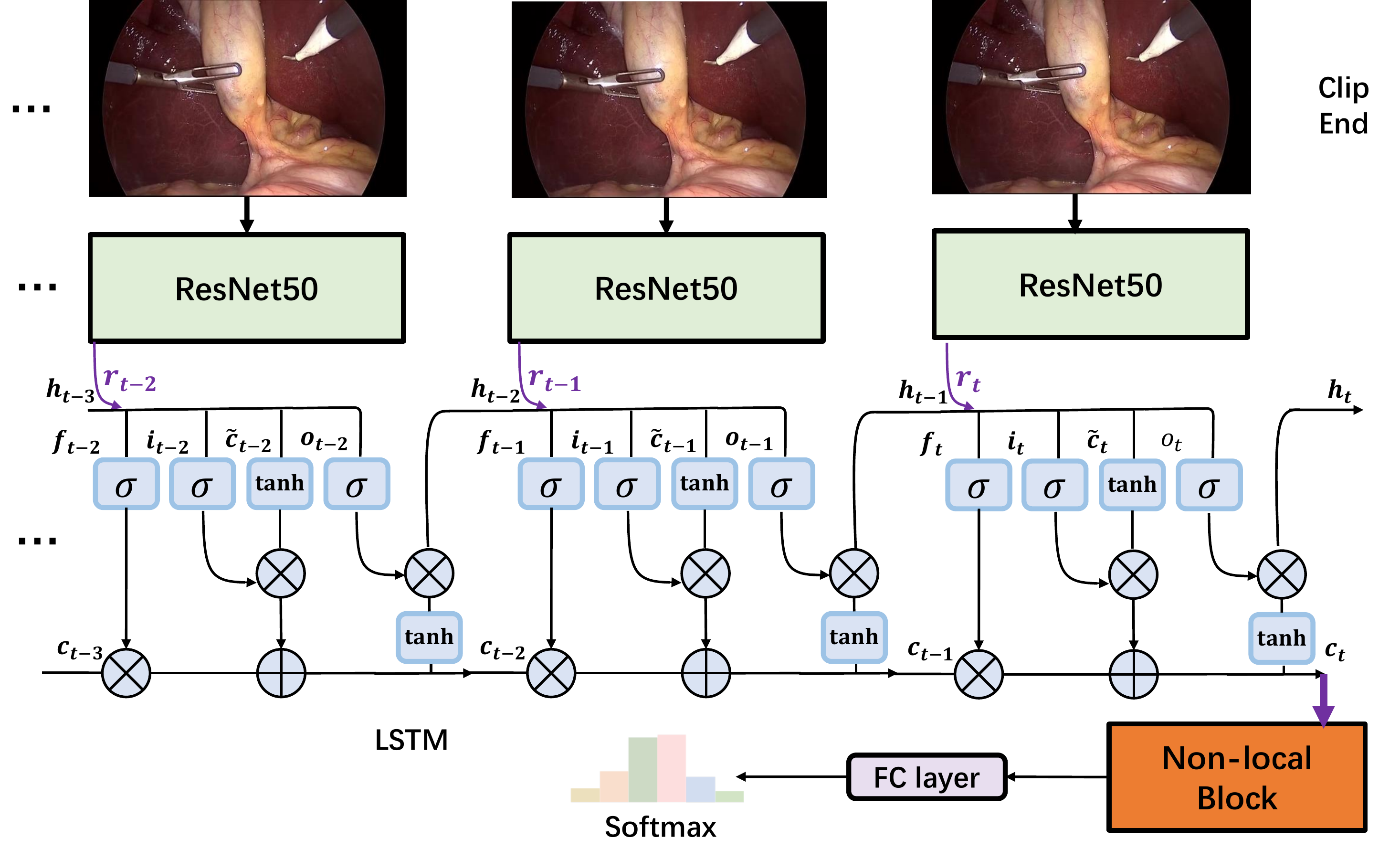}}
	\caption{The overview of our proposed non-local recurrent convolutional network (NL-RCNet) to capture long-range temporal dependency (LRTD) within a video clip for surgical workflow recognition. The output of LSTM unit $c_t$ is flowed to the following non-local block.}
	\label{fig:1}
	\vspace{-7mm}
\end{figure}

In this section, we introduce methods for long-range temporal dependency (LRTD) active learning for surgical workflow recognition task.
Our proposed active learning method is illustrated in Fig.~\ref{fig:1}.
We first train the non-local recurrent convolutional network with the annotated set of $\mathcal{D}_A \!=\! \{(X_{T_q},Y_{T_q})\}_{q=1}^{Q}$, which is initialized with randomly selected $10\%$ data from the unlabelled sample pool $\mathcal{D}_U\!=\! \{X_{T_p}\}_{p=1}^{P}$.
Next, we set up the active learning process by iteratively selecting samples and updating the model.

\subsection{Non-local Recurrent Convolutional Network (NL-RCNet)}
As illustrated in Fig.~\ref{fig:1}, we design a non-local recurrent convolutional network to serve as a foundation for active learning.
To meet the complex surgical environments, we employ the recurrent convolutional network to extract highly discriminative spatio-temporal feature from surgical videos.
We exploit a deep 50-layer residual network (ResNet50)~\cite{he2016deep} to extract high-level visual features from each frame and harness a LSTM network to model the temporal information of sequential frames.
We then seamlessly integrate these two components to form an end-to-end recurrent convolutional network, so that the
complementary information of the visual and temporal features can be sufficiently encoded for more accurate recognition.
Based on this high qualitative feature, we employ the non-local block to capture long-range temporal dependency of frames within each clip.
Different from progressive behavior of convolutional and recurrent operations, non-local operations can directly compute interactions between any two positions in each clip, regardless of their positional distance.
Therefore, it can enhance the feature distinctiveness for better workflow recognition, with the capability of deducing the cross-frame dependency of arbitrary intervals.
Moreover, the non-local block can construct the dependency of each frame in clips with the captured long-range temporal dependency.
Such advantage plays more important roles for our active learning systems, with detailed descriptions in Section~\ref{sample_selection}.

\subsection{Long-Range Temporal Dependency (LRTD) Modeling with Non-local Block}
We introduce the non-local operation for modeling long-range temporal dependency of video clips.
This section describes how we formulate the non-local operation and design a non-local block that can be integrated into the entire framework. Our non-local operation design follows \cite{wang2018non}. 

The non-local operation is designed as follows:
\vspace{-3mm}
\begin{equation}\label{equ1}
\textbf{y}_{t_i} = \frac{1}{\mathcal{C}(\textbf{x})}\sum_{
	\forall j}f(\textbf{x}_{t_i},\textbf{x}_{t_j})g(\textbf{x}_{t_j}).
\vspace{-3mm}
\end{equation}
Here $t_i$ is the index of an output time step whose response is to be computed and $t_j$ is the index that enumerates all possible time steps.
$\textbf{x}$ is the input signal and $\textbf{y}$ is the output signal of the same size as $\textbf{x}$.
Note that $\textbf{x}$ is the high-level spatio-temporal feature outputted from our CNN-LSTM architecture ($\textbf{c}$ in Fig.~\ref{fig:1}), forming a strong base for better non-local dependency modeling.
A pairwise function $f$ computes a scalar between $\textbf{x}_{t_i}$ and all $\textbf{x}_{t_j}$.
The unary function $g$ computes a representation of the input signal at the time step $j$.
The response is normalized by a factor $\mathcal{C}(\textbf{x})$.
The non-local behavior in Eq.~\ref{equ1} is due to the fact that all time steps ($\forall j$) in one clip are considered in the operation.
As a comparison, a recurrent operation only sums up the weighted input from adjacent frames.

\begin{figure}[t!]
	\centering
	\label{figure1c}
	{\includegraphics[width=0.6\linewidth]{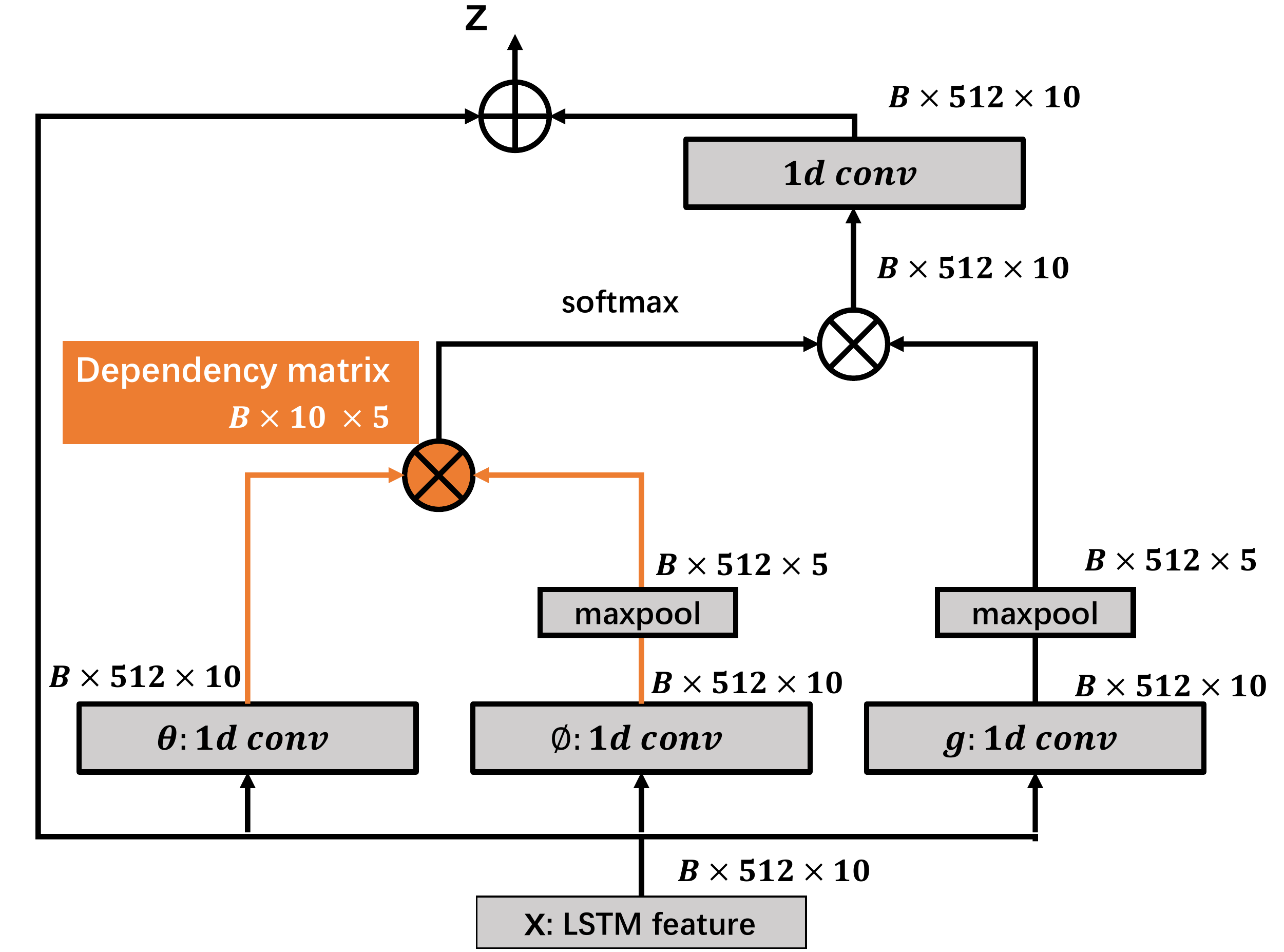}}
	\caption{Non-local block design along time dimension. The intermediate  generated dependency matrix of non-local block can be utilized to represent cross-frame dependency among all the frames in a clip.}
	\label{fig:2}
	\vspace{-5mm}
\end{figure}

Next we describe the calculation of our non-local operator $g$ and $f$. $g$ is defined by a linear embedding: $g(\textbf{x}_{t_j}) = W_g \textbf{x}_{t_j} $, where $W_g$ is the model parameter to be learned.
It is implemented by a $1$D convolution to model the representation in spacetime aspect.
For the definition of function $f$, we choose Embedded Gaussian to compute similarity in an embedding space, where in our case, to compute similarity of embedding features in different time steps:
\begin{equation}\label{equ2}
f(\textbf{x}_{t_i},\textbf{x}_{t_j}) = e^{{\theta(\textbf{x}_{t_i})}^{T}\phi(\textbf{x}_{t_j})},
\end{equation}
where $\theta(\textbf{x}_{t_i}) = W_{\theta}\textbf{x}_{t_i}$ and $\phi(\textbf{x}_{t_j}) = W_{\phi}\textbf{x}_{t_j}$ are two embeddings.
The normalization factor $\mathcal{C}(x)$ in Eq.~\ref{equ1} is set as $\mathcal{C}(x) = \sum_{\forall {t_j}}f(\textbf{x}_{t_i}, \textbf{x}_{t_j})$.

The non-local operation of Eq.~\ref{equ1} is then wrapped into a non-local block, which can be easily incorporated into our CNN-LSTM architecture.
We illustrate the non-local block in Fig.~\ref{fig:2}.
We first obtain the feature $\textbf{x}_{t_i}$ generated by our CNN-LSTM framework.
It is a $B\times512\times10$ matrix ($B$: batch size, $512$: channel number, $10$: clip length), which describes the feature of a 10-second video clip.
Followed Eq.~\ref{equ1}, we calculate $\textbf{y}_{t_i}$ where the pairwise computation of $f(\textbf{x}_{t_i},\textbf{x}_{t_j})$ is done by matrix multiplication as shown in Fig.~\ref{fig:2}.
In our designed non-local block, we then connect the $\textbf{x}_{t_i}$ and $\textbf{y}_{t_i}$ with the residual connection by element-wise addition~\cite{he2016deep}.
Note that the residual connection allows us to insert our non-local block into any pre-trained model, without breaking its initial behavior (e.g., if $W_z$ is initialized as zero).
The overall definition is as follows:
\vspace{-2mm}
\begin{equation}\label{equ3}
\textbf{z}_{t_i} = W_z\textbf{y}_{t_i} + \textbf{x}_{t_i}.
\vspace{-2mm}
\end{equation}

In the practical implementation of the non-local block, we follow the design in~\cite{wang2018non}, and utilize a simple yet effective subsampling strategy to reduce the computation workload when model the dependency among frames.
Concretely, we modify Eq.~\ref{equ1} as: $\textbf{y}_{t_i} = \frac{1}{\mathcal{C}(\hat{\textbf{x}})}\sum_{
	\forall j}f(\textbf{x}_{t_i},\hat{\textbf{x}}_{t_j})g(\hat{\textbf{x}}_{t_j})$,
where $\hat{\textbf{x}}$ is a subsampled version of $\textbf{x}$.
As shown in Fig.~\ref{fig:2}, we add the max pooling layer after $\phi$ and $g$ to achieve this.
Note that this strategy does not alter the non-local behavior, instead, it can make the computation sparser by reducing the amount of pairwise computation of 1/4.
\subsection{Active Sample Selection with Non-local Intra-clip Dependency Score}
\label{sample_selection}
\begin{figure}[t!]
	\centering
	\label{figure1c}
	{\includegraphics[width=1\linewidth]{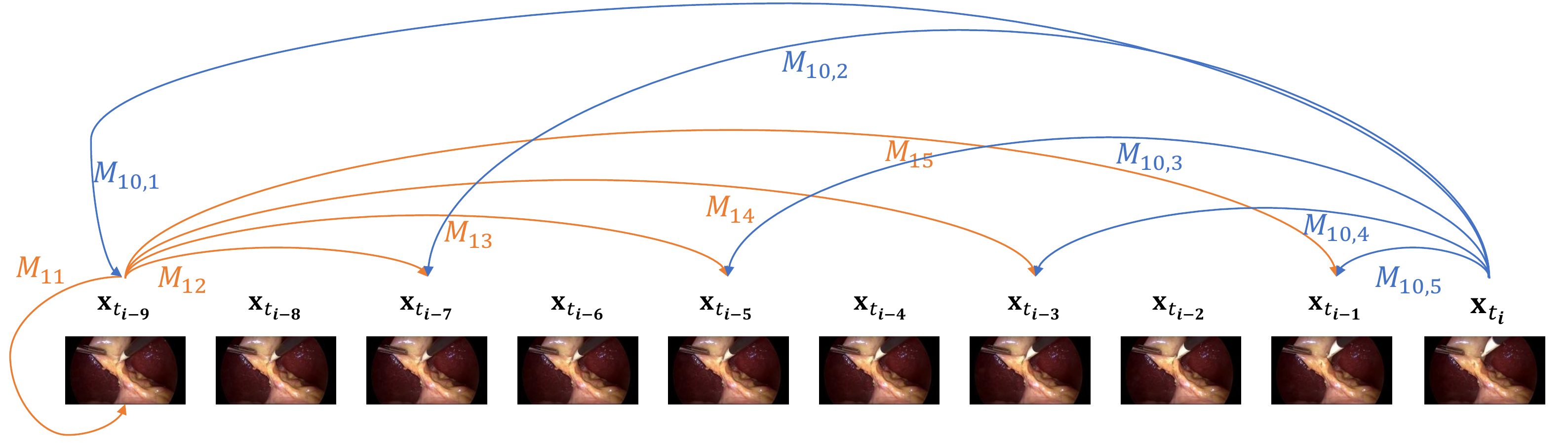}}
	\caption{LRTD based sample selection. LRTD comes from the non-local cross-frame dependency score that is computed by dependency matrix $\mathcal{M}_{mn}$ for clip $X_T$ in Eq.~\ref{eqn:NL-LRTD}. }
	\label{fig:3}
	\vspace{-5mm}
\end{figure}

With using the non-local block, we can obtain the dependency of different frames within one video clip $X_T=\{\textbf{x}_{t_{i-9}},... \textbf{x}_{t_i}\}$.
As illustrated in Fig.~\ref{fig:2}, we get a matrix $\mathcal{M}$ with embedded Gaussian function in Eq.~\ref{equ2}.
Such matrix can represent the intermediate dependency between frames within this clip.
To clearly show the dependency we modeled, we interpret it with Fig.~\ref{fig:3}.
As mentioned before, we use the subsampling strategy in the non-local block to reduce the computational workload.
Moreover, we find that such subsampling on video clip can help to focus the dependency of frames with the intervals, and reduce the effect of neighbouring dependency which has been represented with the LSTM modeling.
To this end, the matrix $\mathcal{M}_{mn}$ for one clip $X_T$ is with $10 \times 5$ dimensions.
For example in Fig.~\ref{fig:3}, $\mathcal{M}_{12}$ is the dependency between the $\textbf{x}_{t_{i-9}}$ and $\textbf{x}_{t_{i-7}}$ with the frame interval of 2.

We select video clips with the weak dependency for annotation query, as they contain richer information and of more representative to better benefit network training.
The model would present relatively weak dependency when the video clip contains some “hard” unlabelled samples, which are usually either severe blur scenes or noise in surgical videos.
In addition, the video clips with high intra-class variance also present weak dependency.
The video clips in both situations are challenging for network to recognize, while in the other hand, demonstrating their high informativeness to train the network more effectively and efficiently.

To select the video clips with the weak dependency, we propose to calculate the overall intra-clip dependency based on the dependency matrix.
For each clip sample $X_T$, we first rank all the values of its dependency matrix $\mathcal{M}_{mn}(X_T)$ in descending order and select the first $N_M$ values.
These values with the strongest dependency responses are verified to better represent the overall dependency of this clip.
We then average these values to obtain a final dependency score $\mathcal{R}$ for each clip sample.
\begin{equation}
\mathcal{R}(X_T) = \frac{1}{N_M} \sum \operatorname{Rank} ( \{ \mathcal{M}_{mn}(X_T) \} , N_M).
\label{eqn:NL-LRTD}
\end{equation}
Given the unlabelled video clip pool $\mathcal{D}_U$, we calculate the intra-clip dependency score for all the clips.
We then rank $\mathcal{D}_U$ according to dependency score, and select the lower ones with weaker dependency and stronger informativeness.
The selected clip samples following this criteria are represented as $\mathcal{S}_{C}$:
\begin{equation}
{\mathcal{S}_{C}} \gets \underset{X_T}{\operatorname{Rank}}( \{ \mathcal{R}(X_T) \}, N_C),
\label{eqn:rank-LRTD}
\end{equation}
where $\mathcal{R}(X_T)$ is dependency score for each clip $X_T$ in $\mathcal{D}_U$, ranking is in ascending order, and the first $N_C$ samples are selected.
We set $N_M = 5$ and $N_C = 10\% \times N$ where $N$ is the total number of available clips.
5 is the hyper-parameter which controls the degree when representing intra-clip dependency. It is set based on the dimension of dependency matrix $\mathcal{M}_{mn}$.
$10\%$ is to control the scale of newly selected clips in each round of sample selection, which follows the traditional design of active learning based methods \cite{yang2017suggestive,zhou2017fine,shi2019active}.

\begin{table}[h!]
	\centering
	\caption{The network architecture of our NL-RCNet model. From Conv1\_x to Conv5\_x, we follow the 50-layer residual network design, then we use LSTM to capture temporal information and non-local block to capture intra-clip dependency.}
	\label{tab:res}
	\resizebox{0.6\textwidth}{!}{
		\begin{tabular}{ccc}
			\toprule
			\multicolumn{1}{c|}{\textbf{Layer name}}
			& \multicolumn{1}{c|}{\textbf{Output size}}
			& \multicolumn{1}{c}{\textbf{NL-RCNet}}
			\\
			\midrule
			\multicolumn{1}{c|}{Conv1\_x}
			& \multicolumn{1}{c|}{$112\times112$}
			& \multicolumn{1}{c}{$7\times7$, 64, stride2}
			\\
			\hline
			\multicolumn{1}{c|}{\multirow{2}{*}{Conv2\_x}}
			& \multicolumn{1}{c|}{\multirow{2}{*}{$56\times56$}}
			& \multicolumn{1}{c}{$3\times3$ max pool, stride 2}
			\\
			\cline{3-3}
			\multicolumn{1}{c|}{}
			& \multicolumn{1}{c|}{}
			& \multicolumn{1}{c}{$\begin{bmatrix}  1 \times 1, 64  \\ 3 \times 3, 64  \\ 1 \times 1, 256  \end{bmatrix} \times 3$}            \\ \hline
			\multicolumn{1}{c|}{Conv3\_x}                  & \multicolumn{1}{c|}{$28\times28$}                  & \multicolumn{1}{c}{$\begin{bmatrix}  1 \times 1, 128  \\ 3 \times 3, 128  \\ 1 \times 1, 512  \end{bmatrix} \times 4$}            \\ \hline
			\multicolumn{1}{c|}{Conv4\_x}                  & \multicolumn{1}{c|}{$14\times14$}                  & \multicolumn{1}{c}{$\begin{bmatrix}  1 \times 1, 256  \\ 3 \times 3, 256  \\ 1 \times 1, 1024  \end{bmatrix} \times 23$}           \\ \hline
			\multicolumn{1}{c|}{Conv5\_x}                  & \multicolumn{1}{c|}{$7\times7$}                    & \multicolumn{1}{c}{$\begin{bmatrix}  1 \times 1, 512  \\ 3 \times 3, 512 \\ 1 \times 1, 2048  \end{bmatrix} \times 3$}
			\\ \hline
			\multicolumn{1}{c|}{Average Pool}                  & \multicolumn{1}{c|}{$1\times1$}                    & \multicolumn{1}{c}{$\begin{bmatrix}7 \times 7, 2048  \end{bmatrix} \times 1$}            \\
			\hline
			\multicolumn{1}{c|}{LSTM}                  & \multicolumn{1}{c|}{$1\times1$}                    & \multicolumn{1}{c}{$\begin{bmatrix}  10, 512  \end{bmatrix} \times 1$} \\
			\hline
			\multicolumn{1}{c|}{Non-local block}                  & \multicolumn{1}{c|}{$1\times1$}                    & \multicolumn{1}{c}{Fig.~\ref{fig:2}}
			\\ \hline
			\multicolumn{1}{c|}{Output}                    & \multicolumn{1}{c|}{$1\times1$}                    & \multicolumn{1}{c}{max pool, fc, softmax} \\
			\bottomrule
			\multicolumn{1}{l}{}                            & \multicolumn{1}{l}{}                        & \multicolumn{1}{l}{}                               \\
			\multicolumn{1}{l}{}                            & \multicolumn{1}{l}{}                        & \multicolumn{1}{l}{}                               \\
			\multicolumn{1}{l}{}                            & \multicolumn{1}{l}{}                        & \multicolumn{1}{l}{}                               \\
			\multicolumn{1}{l}{}                            & \multicolumn{1}{l}{}                        & \multicolumn{1}{l}{}                               \\
			\multicolumn{1}{l}{}                            & \multicolumn{1}{l}{}                        & \multicolumn{1}{l}{}                               \\
			\multicolumn{1}{l}{}                            & \multicolumn{1}{l}{}                        & \multicolumn{1}{l}{}
		\end{tabular}
	}
\vspace{-16mm}
\end{table}

\subsection{Implementation Details of our Active Learning Approach}
We first train the recurrent convolutional network with the annotated set of $\mathcal{D}_A \!=\! \{(X_{T_q},Y_{T_q})\}_{q=1}^{Q}$,
which is initialized with randomly selected 10\% data from the unlabelled sample pool $\mathcal{D}_U\!=\! \{X_{T_p}\}_{p=1}^{P}$.
We then train the entire NL-RCNet in an end-to-end manner with the parameters of recurrent convolutional part initialized by the pre-trained model, and non-local block is randomly initialized.
The whole network architecture is illustrated using Table~\ref{tab:res}.

Next, we start our active learning process using previous backbone, i.e. NL-RCNet.
We iteratively select samples by LRTD method and update $\mathcal{D}_A$.
By jointly training with newly added annotated data of $\mathcal{D}_A$, we progressively update the model.
In each update, we first pre-train the recurrent convolutional part (CNN-LSTM model) to learn reliable parameters for the following initialization in the overall network and here we initialize the ResNet50 with weights trained on the ImageNet dataset ~\cite{he2016deep}.
We use back-propagation with stochastic gradient descent to train the model.
The learning rates of CNN module and LSTM module are initialized by $5 \! \times \! 10^{-5}$ and $5 \! \times \! 10^{-4}$, respectively. Both of them are divided by a factor of 10  every 3 epochs.
After obtaining the pre-trained CNN-LSTM model, we train the entire NL-RCNet in an end-to-end manner.
The network is fine-tuned by Adam optimizer, where learning rate of CNN-LSTM part and non-local block are initialized by $5 \! \times \! 10^{-5}$ and $5 \! \times \! 10^{-4}$, and are also reduced by 10 every 3 epochs.
The loss functions for CNN-LSTM and NL-RCNet are both cross entropy losses and stop with 25-epoch training.
As for input process, 
we resize the frames from the original resolution of $1920\times1080$ and $854\times480$ into $250\times250$ to dramatically save memory and reduce network parameters.
In order to enlarge the training dataset, we apply automatic augmentation with random $224\times224$ cropping, horizontal flips by a factor of $0.5$, random rotations of $[-10,10]$ degrees, brightness, saturation and contrasts by a random factor of $0.2$, and hue by a random factor of $0.05$.
Our framework is implemented based on the PyTorch using 4 GPUs for acceleration.
\section{Experiment}
\subsection{Dataset and Evaluation Metrics}
We extensively validate our LRTD based active learning method on a popular public surgical dataset Cholec80~\cite{twinanda2016endonet}.
The dataset consists of 80 videos recording the cholecystectomy procedures performed by 13 surgeons.
The videos are captured at 25 fps and each frame has the resolution of $854 \times 480$ or $1920 \times 1080$.
All the frames are labelled with 7 defined phases by experts.
For fair comparison, we follow the same evaluation procedure reported in~\cite{twinanda2016endonet}, splitting the dataset into two subsets with equal size, with 40 videos for training and the rest 40 videos for testing. For data generalization strategy, we create each clip sequentially in the form of a sliding window, with each time shifting one frame forward, which means 9 frames overlap between two continuous clips. So does the test time clip generation strategy. Moreover, one clip-wise annotation corresponding one frame-wise because we only utilize the last frame's annotation during training.
We conduct all the experiments in the online mode, by only using the preceding frames for recognition.
The computing time for selection between two annotation stages is 0.58s/clip on a workstation with 1 Nvidia TITAN Xp.

To quantitatively analyze the performance of our method, we employ four metrics to evaluate our methods, including Accuracy(ACC), Precision (PR), Recall (RE), Jaccard (JA) and F1 Score (F1).
PR, RE, JA and F1 Score are computed in phase-wise, defined as:
\begin{equation}
\label{eq:eva}
\centering
\begin{gathered}
\mathrm{PR}=\frac{|\mathrm{GT} \cap \mathrm{P}|}{|\mathrm{P}|}, ~ \mathrm{RE}=\frac{|\mathrm{GT} \cap \mathrm{P}|}{|\mathrm{GT}|}, ~
\mathrm{JA}=\frac{|\mathrm{GT} \cap \mathrm{P}|}{|\mathrm{GT} \cup \mathrm{P}|}, ~
\mathrm{F1}=\frac{2}{\frac{1}{\mathrm{PR} } + \frac{1}{\mathrm{RE}}}, \\
\end{gathered}
\small
\end{equation}
where $\mathrm{GT}$ and $\mathrm{P}$ represent the ground truth set and prediction set of one phase, respectively.
After PR, RE, JA and F1 of each phase are calculated, we average these values over all the phases and obtain them of the entire video.
The ACC is calculated at video-level, defined as the percentage of frames correctly classified into the ground truths in the entire video.
\vspace{-5mm}
\begin{table}
	\centering
	\caption{Surgical workflow recognition performance of different methods under settings of full annotation and active learning (mean$\pm$std., \%).
	}
	\resizebox{1\textwidth}{!}
	{
		\begin{tabular}{|c|l|c|c|c|c|c|c|}
			\hline
			\multicolumn{2}{|c|}{\begin{tabular}[c]{@{}c@{}}Methods\\\end{tabular}} & \begin{tabular}[c]{@{}c@{}}Data~ \\Amount\end{tabular} & \begin{tabular}[c]{@{}c@{}}Accuracy\\\end{tabular} & \begin{tabular}[c]{@{}c@{}}Precision\\\end{tabular} & \begin{tabular}[c]{@{}c@{}}Recall\\\end{tabular} &
			\begin{tabular}[c]{@{}c@{}}Jaccard \\\end{tabular}&
			\begin{tabular}[c]{@{}c@{}}F1 Score	\\\end{tabular}  \\
			\hline
			\hline
			\multirow{2}{*}{Full}
			& EndoNet~\cite{twinanda2016endonet}
			& 100\%
			& ~$81.70\pm{4.20}$~
			& ~$73.70\pm{16.10}$~
			& ~$79.60\pm{7.90}$~
			& ~$-$~
			& ~$-$~
			\\
			\cline{2-8}
			\multirow{2}{*}{Annotation}	
			& SV-RCNet~\cite{jin2017sv}
			& 100\%
			& ~\bm{$86.40\pm{7.30}$}~
			& ~$82.90\pm{5.90}$~
			& ~$84.50\pm{8.00}$~
			& ~$-$~
			& ~$-$~
			\\
			\cline{2-8}
			& NL-RCNet \textbf{(Ours)  }
			& 100\%
			& ~$85.73\pm{6.96}$~
			& ~\bm{$82.94\pm{6.20}$}~
			& ~\bm{$85.04\pm{5.15}$}~
			& ~$69.96\pm{8.83}$~
			& ~$82.08\pm{6.45}$~
			\\
			\hline
			\hline
			\multirow{2}{*}{Active} & Full Data
			& 100\%
			& ~$85.73\pm{6.96}$~
			& ~$82.94\pm{6.20}$~
			& ~$85.04\pm{5.15}$~
			& ~\bm{$69.96\pm{8.83}$}~
			& ~$82.08\pm{6.45}$~
			\\
			\cline{2-8}
			\multirow{2}{*}{Learning} & DBN~\cite{bodenstedt2019active}
			& 50\%
			& ~$84.88\pm{8.35}$~
			& ~\bm{$83.00\pm{6.16}$}~
			& ~$83.54\pm{5.73}$~
			& ~$69.04\pm{7.59}$~
			& ~$81.34\pm{5.46}$~
			\\
			\cline{2-8}
			& CNNLSTM-EMB
			& 50\%
			& ~$85.13\pm{8.10}$~
			& ~$82.16\pm{7.09}$~
			& ~$84.10\pm{5.26}$~
			& ~$68.72\pm{8.69}$~
			& ~$81.09\pm{6.26}$~
			\\
			\cline{2-8}
			& LRTD \textbf{(Ours)}
			& 50\%
			& ~\bm{$85.87\pm{7.36}$}~
			& ~$82.76\pm{6.85}$~
			& ~\bm{$85.26\pm{4.30}$}~
			& ~$69.94\pm{8.99}$~
			& ~\bm{$82.13\pm{6.22}$}~
			\\
			\hline
		\end{tabular}
	}
	\label{table:1}
\end{table}
\vspace{-5mm}
\subsection{Quantitative Results and Comparison with Other Methods}
In our active learning process, based on the initially randomly selected 10\% data,
we select and iteratively add training samples until obtaining predictions which cannot be significantly improved ($p>0.05$) over the accuracy of last round.
It turns out that we only need 50\% of the data for workflow recognition task.

\begin{figure}[h]
	\centering
	{\includegraphics[width=1\linewidth]{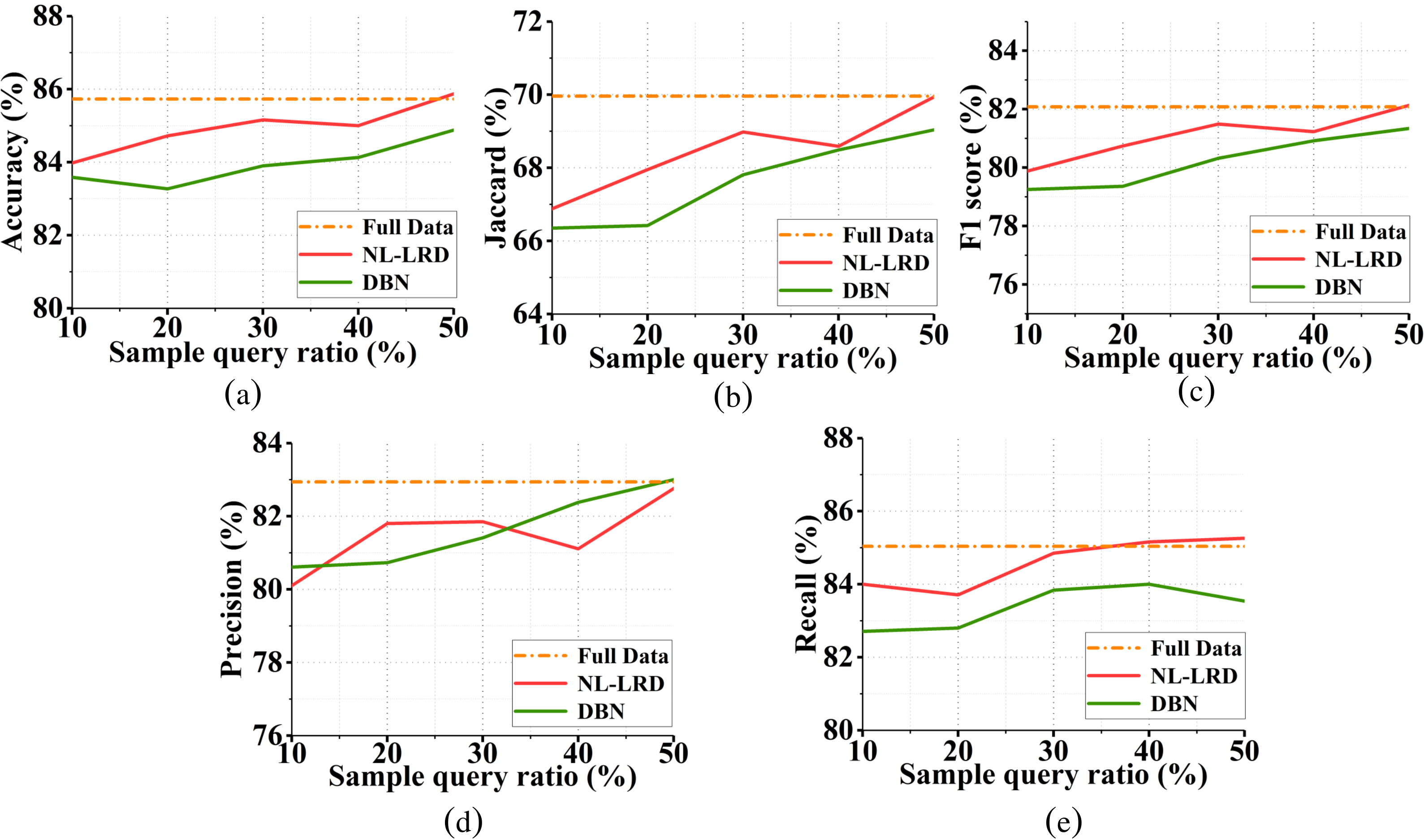}}
	\caption{Comparison of our proposed LRTD method with state-of-the-art DBN~\cite{bodenstedt2019active} for active learning on various metrics of (a) Accuracy, (b) Jaccard, (c) F1 Score, (d) Precision and (e) Recall.}
	\label{fig:experiment}
\end{figure}
In Table~\ref{table:1}, we divide the different comparison methods into two groups, i.e., the fully supervised methods in this workflow recognition task, and active learning only with sample selection strategy.
The amount of employed annotated data is indicated in data amount column.
For the fully supervised comparison, we include two state-of-the-art methods, i.e. EndoNet~\cite{twinanda2016endonet} and SV-RCNet~\cite{jin2017sv}.
We observe that our NL-RCNet can slightly outperform these two methods, with adding the non-local block to capture the long-range temporal dependency.

Moreover, the more important point of adding this block is for active learning part, by using the cross-frame dependency matrix, which is the intermediate result of this block.
We implement full-data training as standard bound, and we compare with the state-of-the-art active learning method for workflow recognition~\cite{bodenstedt2019active}, which use the deep Bayesian network (DBN) to estimate uncertainty for sample selection.
Note that \cite{bodenstedt2019active} does not follow the common train-test data split setting. Therefore, we re-implement this method by using the same evaluation process and the same ResNet50-LSTM network architecture for the fair comparison.
From Table~\ref{table:1}, we see that our LRTD based strategy achieves better performance than DBN method in 50\% data ratio, in particular, improving around 1\% for Accuracy.
To verify the effectiveness of our non-local block to capture the dependency, we also conduct an ablation study named CNNLSTM-EMB. It uses pure CNN-LSTM without non-local block to train the network, and the dependency matrix is calculated using the dot-product similarity on the frame embeddings output by the CNN-LSTM network. We can see that our LRTD achieves the superior performance to CNNLSTM-EMB in all the evaluation metrics, demonstrating that the non-local block can better construct the intra-clip dependency.

In addition, we can see that our method reaches state-of-the-art performance and even surpass results against full-data training with significantly cost-effective labellings (i.e. only $50\%$ annotation).
Specially, our LRTD based active learning achieves slightly better F1 Score performance than the network with full-data training.
Because our LRTD based selection is not only consider the information from previous adjacent frame, but also consider the cross-frame dependency among a whole clip with 10 seconds length.
By modeling the long-range temporal dependency in time dimension, this strategy encourages the prediction more consistent and robust.

We further conduct statistical test by calculating the p-values to compare the state-of-the-art results and our method, with the numbers $2.136\times10^{-14}$ for DBN and our LRTD, 0.044 for Full Data training and LRTD. We get both $p < 0.05$, which indicates a significant improvement for our approach. 
Moreover, we repeat experiments of NL-RCNet(ours), CNNLSTM-EMB, DBN and LRTD(ours) with doing the initial labeled data selection randomly, to verify that the result improvement is encouraged by the effectiveness of our methods. 
The p-values for NL-RCNet(ours) and LRTD(ours) are 0.18 and 0.71, respectively. 
Both of them are larger than 0.05, indicating that the two rounds’ results do not have much significance. 
While the p-value of DBN and CNNLSTM-EMB are separately $2.90\times10^{-7}$ and $0.002$, which are smaller than 0.05, demonstrating that the two-round results of DBN or CNNLSTM-EMB have relatively large gap. The underlying reason is that DBN is sensitive to initially selected labeled data when conducting active learning process, so it is not as robust and stable as our LRTD strategy. For CNNLSTM-EMB, the relation matrix shows less effectiveness than LRTD, which causes the fluctuation on data representativeness among those 50\% selected data in two runs, so the performance is not stable.

\begin{figure}[h!]
	\centering
	\includegraphics[width=0.9\linewidth]{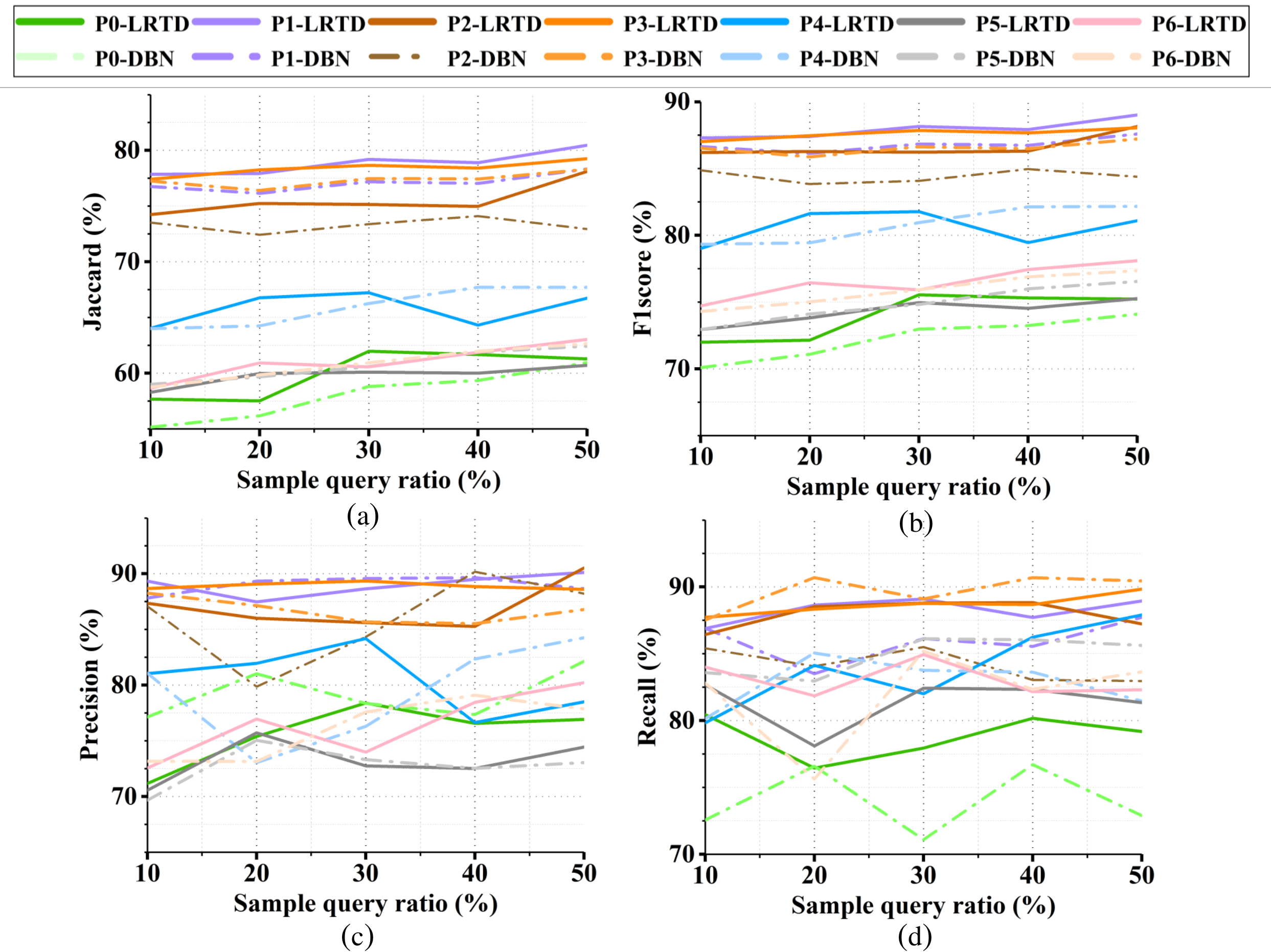}
	\vspace{-3mm}
	\caption{Comparison of our proposed LRTD method with state-of-the-art DBN~\cite{bodenstedt2019active} for active learning on various metrics of Jaccard, F1 Score, Precision and Recall, with results on each phase at different annotation ratios. P0-P6 separately corresponds to each surgical phase in our dataset, and also corresponds to each phase name given in Table~\ref{phaseresults} (Appendix).}
	\label{fig:phaseresults}
\end{figure}
\begin{figure}[h!]
	\centering
	\vspace{-3mm}
	\subcaptionbox{\label{select}}{\includegraphics[width=1\linewidth]{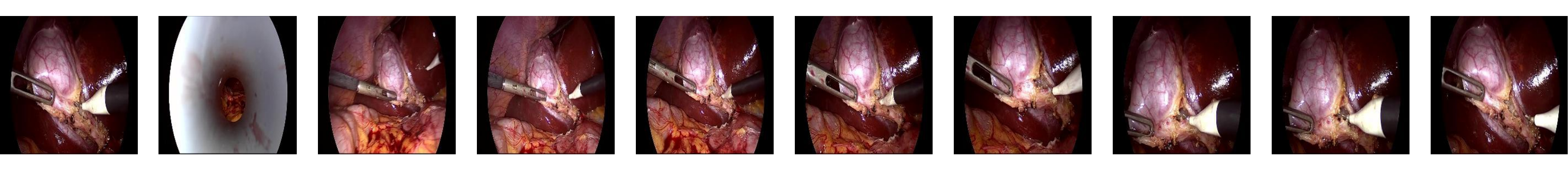}}
	\subcaptionbox{\label{notselect}}{\includegraphics[width=1\linewidth]{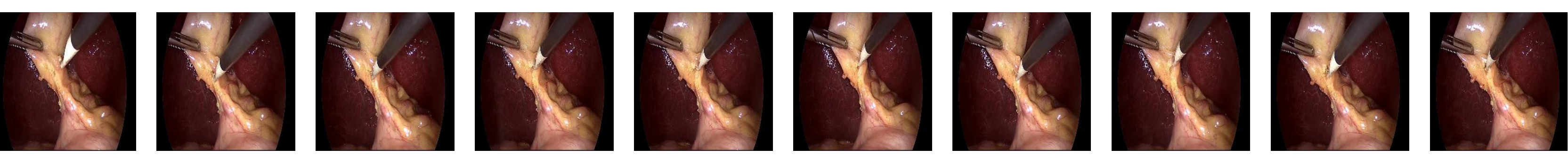}}
	\includegraphics[width=0.8\linewidth]{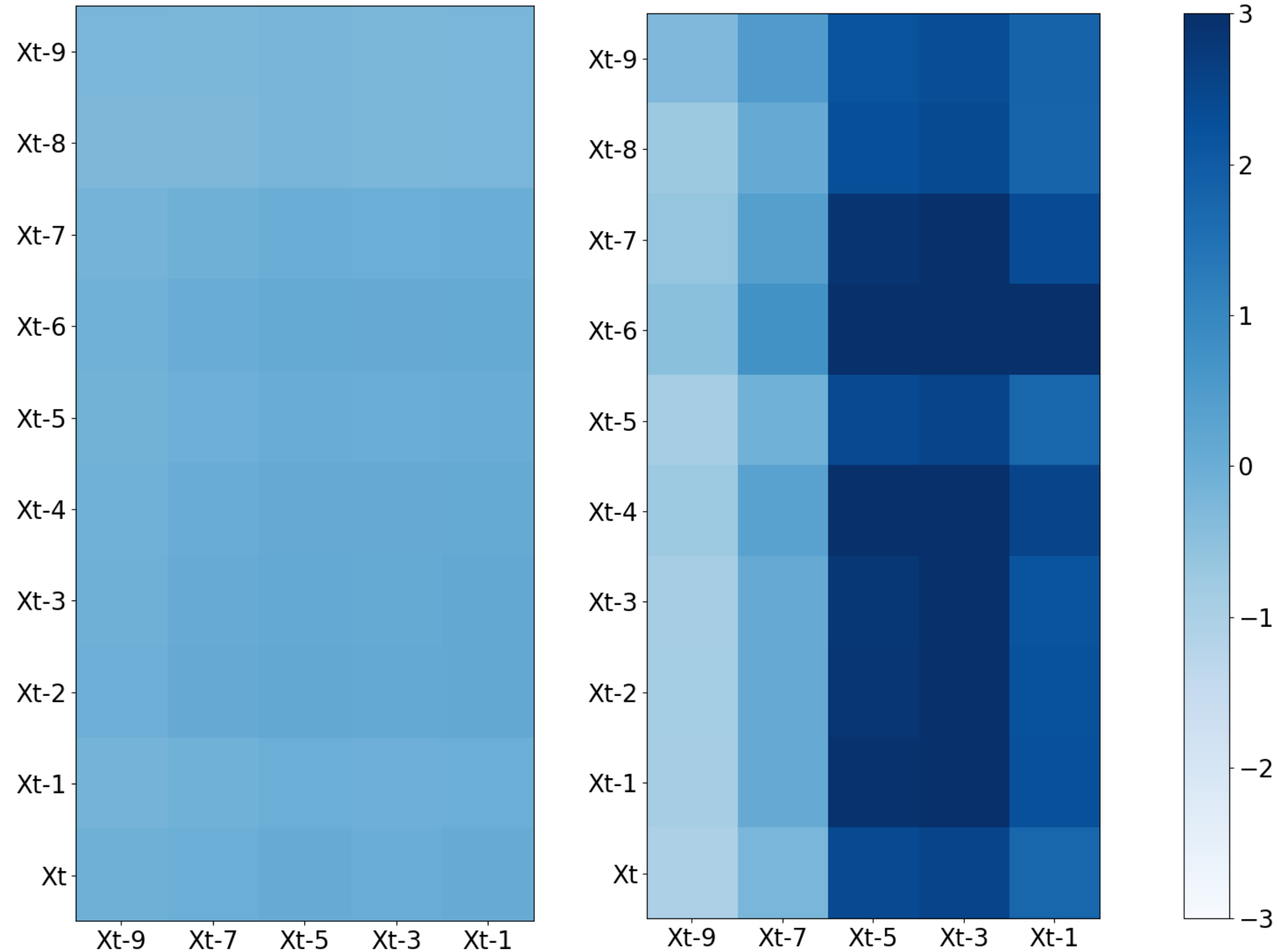}
	\vspace{-3mm}
	\caption{(a) One selected clip sample with weak intra-clip dependency, so the color brightness of its dependency matrix in (c) is low in most matrix positions; (b) one unselected clip sample with strong intra-clip dependency, so the color brightness of its dependency matrix in (d) are high in most matrix positions; (c)-(d) the visualization of corresponding dependency matrices of clip (a) and (b).}
	\label{fig:selectnoselect}
	\vspace{-5mm}
\end{figure}
\subsection{Analytical Experiments of Our Proposed LRTD Approach}
For detailed analysis of our LRTD method compared with other active learning method, we conduct the sub-experiments in each sample ratio, with totally five groups until 50\% annotations.
The quantitative comparison results are listed in Table~\ref{table:2} (see Appendix).
We can see that the performance of our LRTD gradually increase through different sample ratios, and can keeps higher than results of DBN in terms of Accuracy, Jaccard, F1 Score and Recall.
To more clearly show the change tendency of results with data gradually added by different selection strategies for network training, we draw the changing curves in Fig.~\ref{fig:experiment}.
We can see that both LRTD and DBN can stably promote Accuracy, Jaccard, and F1 Score performance without a huge fluctuation until the p-value larger than $0.05$ ($50\%$ data ratio).
However, DBN shows fluctuation in Recall while LRTD shows fluctuation on Precision, as both of them do not consider data diversity when selecting samples.
Therefore some newly selected data would change the data distribution of training set and result in unstable performance.
We further present Fig.~\ref{fig:phaseresults} to provide more details to show the results in each phase-level through different annotation ratios (quantitative result details can be found in Table~\ref{phaseresults} (see Appendix)).
It is observed that LRTD consistently improves the performance in almost phases with the increasing annotation ratio.
Compared with DBN, our method achieves better results in Phase 0-3 while DBN performs better in Phase 4-6 in various annotation ratios.

To intuitively show the long-range temporal dependency across frames, and provide the insight of why we choose the clips with weak dependencies to annotate, we illustrate one selected clip and one unselected clip using our LRTD method in Fig.~\ref{fig:selectnoselect}.
From selected clip in Fig.~\ref{select}, we find they have low dependency in a long-range temporal dependency thus cause the cross-frame dependency score quite low.
This can be clearly seen by Fig.~\ref{fig:selectnoselect}\textcolor{blue}{c}, that the color brightness is low in many positions.
Such sample is informative to our model.
However, in Fig.~\ref{notselect}, we can find that these frames are highly related to each other, so the dependency scores are high with the strong brightness (see Fig.~\ref{fig:selectnoselect}\textcolor{blue}{d}), which has low information for the model to train.
We further analyze which phases occupy more important proportion in the selected clips and illustrate the percentage in Fig.~\ref{fig:histogram} (see Appendix). It is observed that P1 (43.0\%) and P4 (27.5\%) surpass other phases in ratio value, while clips containing phase transition only occupy 2.2\%.  It is reasonable as the phase proportion of the selected clips is corresponding to the original training data, where P1 and P3 take relatively longer duration in the surgical procedure.
\vspace{-3mm}
\section{Conclusion}

In this paper, we propose a long-range temporal dependency (LRTD) based active learning for surgical workflow recognition.
By modeling the cross-frame dependency within video clips, we select clips with weaker dependency for annotation query.
Our network achieves superior performance of workflow recognition over other state-of-the-art active learning methods on a popular public surgical dataset.
By only requiring labeling 50\% clips, our method can surpass fully-supervised counterpart, which endorses the potential value in clinical practice.\\
\textbf{Conflict of interest} Xueying Shi, Yueming Jin, Qi Dou and Pheng-Ann Heng declare that they have no conflict of interest.\\
\textbf{Ethical approval} For this type of study formal consent is not required.\\
\textbf{Informed consent} This article contains patient data from publicly available datasets.
\\
\\
\textbf{Acknowledgments.} 
The work was partially supported by HK RGC TRS project
T42-409/18-R, and a grant from the National Natural Science Foundation of China
(Project No. U1813204) and CUHK T Stone Robotics Institute.
\vspace{-3mm}
\bibliographystyle{abbrv}
\bibliography{ref}

\begin{thebibliography}{10}

\bibitem{ahmidi2017dataset1}
N.~Ahmidi, L.~Tao, S.~Sefati, Y.~Gao, C.~Lea, B.~B. Haro, L.~Zappella,
  S.~Khudanpur, R.~Vidal, and G.~D. Hager.
\newblock A dataset and benchmarks for segmentation and recognition of gestures
  in robotic surgery.
\newblock {\em IEEE Transactions on Biomedical Engineering}, 64(9):2025--2041,
  2017.

\bibitem{bodenstedt2019active}
S.~Bodenstedt, D.~Rivoir, A.~Jenke, M.~Wagner, M.~Breucha, B.~M{\"u}ller-Stich,
  S.~T. Mees, J.~Weitz, and S.~Speidel.
\newblock Active learning using deep \protect{Bayesian} networks for surgical
  workflow analysis.
\newblock {\em International Journal of Computer Assisted Radiology and
  Surgery}, 14(6):1079--1087, 2019.

\bibitem{bodenstedt2017unsupervised}
S.~Bodenstedt, M.~Wagner, D.~Kati{\'c}, P.~Mietkowski, B.~Mayer, H.~Kenngott,
  B.~M{\"u}ller-Stich, R.~Dillmann, and S.~Speidel.
\newblock Unsupervised temporal context learning using convolutional neural
  networks for laparoscopic workflow analysis.
\newblock {\em arXiv preprint arXiv:1702.03684}, 2017.

\bibitem{bouget2017vision}
D.~Bouget, M.~Allan, D.~Stoyanov, and P.~Jannin.
\newblock Vision-based and marker-less surgical tool detection and tracking: a
  review of the literature.
\newblock {\em Medical Image Analysis}, 35:633--654, 2017.

\bibitem{bouget2015detecting}
D.~Bouget, R.~Benenson, M.~Omran, L.~Riffaud, B.~Schiele, and P.~Jannin.
\newblock Detecting surgical tools by modelling local appearance and global
  shape.
\newblock {\em IEEE Transactions on Medical Imaging}, 34(12):2603--2617, 2015.

\bibitem{bricon2007context}
N.~Bricon-Souf and C.~R. Newman.
\newblock Context awareness in health care: A review.
\newblock {\em International Journal of Medical Informatics}, 76(1):2--12,
  2007.

\bibitem{cleary2005or}
K.~Cleary and A.~Kinsella.
\newblock \protect{OR} 2020: the operating room of the future.
\newblock {\em Journal of laparoscopic \& advanced surgical techniques. Part
  A}, 15(5):495--497, 2005.

\bibitem{dergachyova2016automatic}
O.~Dergachyova, D.~Bouget, A.~Huaulm{\'e}, X.~Morandi, and P.~Jannin.
\newblock Automatic data-driven real-time segmentation and recognition of
  surgical workflow.
\newblock {\em International Journal of Computer Assisted Radiology and
  Surgery}, 11(6):1081--1089, 2016.

\bibitem{doersch2017multi}
C.~Doersch and A.~Zisserman.
\newblock Multi-task self-supervised visual learning.
\newblock In {\em IEEE International Conference on Computer Vision}, pages
  2051--2060, 2017.

\bibitem{forestier2015automatic}
G.~Forestier, L.~Riffaud, and P.~Jannin.
\newblock Automatic phase prediction from low-level surgical activities.
\newblock {\em International Journal of Computer Assisted Radiology and
  Surgery}, 10(6):833--841, 2015.

\bibitem{funke2018temporal}
I.~Funke, A.~Jenke, S.~T. Mees, J.~Weitz, S.~Speidel, and S.~Bodenstedt.
\newblock Temporal coherence-based self-supervised learning for laparoscopic
  workflow analysis.
\newblock In {\em OR 2.0 Context-Aware Operating Theaters, Computer Assisted
  Robotic Endoscopy, Clinical Image-Based Procedures, and Skin Image Analysis},
  pages 85--93. Springer, 2018.

\bibitem{he2016deep}
K.~He, X.~Zhang, S.~Ren, and J.~Sun.
\newblock Deep residual learning for image recognition.
\newblock In {\em IEEE Conference on Computer Vision and Pattern Recognition},
  pages 770--778, 2016.

\bibitem{james2007eye}
A.~James, D.~Vieira, B.~Lo, A.~Darzi, and G.-Z. Yang.
\newblock Eye-gaze driven surgical workflow segmentation.
\newblock In {\em International Conference on Medical Image Computing and
  Computer-Assisted Intervention}, pages 110--117. Springer, 2007.

\bibitem{jin2017sv}
Y.~Jin, Q.~Dou, H.~Chen, L.~Yu, J.~Qin, C.-W. Fu, and P.-A. Heng.
\newblock \protect{SV-RCNet}: workflow recognition from surgical videos using
  recurrent convolutional network.
\newblock {\em IEEE Transactions on Medical Imaging}, 37(5):1114--1126, 2017.

\bibitem{jin2019multi}
Y.~Jin, H.~Li, Q.~Dou, H.~Chen, J.~Qin, C.-W. Fu, and P.-A. Heng.
\newblock Multi-task recurrent convolutional network with correlation loss for
  surgical video analysis.
\newblock {\em Medical Image Analysis}, page 101572, 2019.

\bibitem{mahapatra2018efficient}
D.~Mahapatra, B.~Bozorgtabar, J.-P. Thiran, and M.~Reyes.
\newblock Efficient active learning for image classification and segmentation
  using a sample selection and conditional generative adversarial network.
\newblock In {\em International Conference on Medical Image Computing and
  Computer-Assisted Intervention}, pages 580--588, 2018.

\bibitem{quellec2014real}
G.~Quellec, K.~Charri{\`e}re, M.~Lamard, Z.~Droueche, C.~Roux, B.~Cochener, and
  G.~Cazuguel.
\newblock Real-time recognition of surgical tasks in eye surgery videos.
\newblock {\em Medical Image Analysis}, 18(3):579--590, 2014.

\bibitem{ross2018exploiting}
T.~Ross, D.~Zimmerer, A.~Vemuri, F.~Isensee, M.~Wiesenfarth, S.~Bodenstedt,
  F.~Both, P.~Kessler, M.~Wagner, B.~M{\"u}ller, H.~Kenngott, S.~Speidel,
  A.~Kopp-Schneider, K.~Maier-Hein, and L.~Maier-Hein.
\newblock Exploiting the potential of unlabeled endoscopic video data with
  self-supervised learning.
\newblock {\em International Journal of Computer Assisted Radiology and
  Surgery}, 13(6):925--933, 2018.

\bibitem{settles.tr09}
B.~Settles.
\newblock Active learning literature survey.
\newblock Computer Sciences Technical Report 1648, University of
  Wisconsin--Madison, 2009.

\bibitem{shi2019active}
X.~Shi, Q.~Dou, C.~Xue, J.~Qin, H.~Chen, and P.-A. Heng.
\newblock An active learning approach for reducing annotation cost in skin
  lesion analysis.
\newblock In {\em International Workshop on Machine Learning in Medical
  Imaging}, pages 628--636. Springer, 2019.

\bibitem{twinanda2016endonet}
A.~P. Twinanda, S.~Shehata, D.~Mutter, J.~Marescaux, M.~De~Mathelin, and
  N.~Padoy.
\newblock Endonet: a deep architecture for recognition tasks on laparoscopic
  videos.
\newblock {\em IEEE Transactions on Medical Imaging}, 36(1):86--97, 2016.

\bibitem{wang2018non}
X.~Wang, R.~Girshick, A.~Gupta, and K.~He.
\newblock Non-local neural networks.
\newblock In {\em IEEE Conference on Computer Vision and Pattern Recognition},
  pages 7794--7803, 2018.

\bibitem{yang2017suggestive}
L.~Yang, Y.~Zhang, J.~Chen, S.~Zhang, and D.~Z. Chen.
\newblock Suggestive annotation: A deep active learning framework for
  biomedical image segmentation.
\newblock In {\em International Conference on Medical Image Computing and
  Computer-Assisted Intervention}, pages 399--407, 2017.

\bibitem{yengera2018less}
G.~Yengera, D.~Mutter, J.~Marescaux, and N.~Padoy.
\newblock Less is more: surgical phase recognition with less annotations
  through self-supervised pre-training of \protect{CNN-LSTM} networks.
\newblock {\em arXiv preprint arXiv:1805.08569}, 2018.

\bibitem{yu2018learning}
T.~Yu, D.~Mutter, J.~Marescaux, and N.~Padoy.
\newblock Learning from a tiny dataset of manual annotations: a teacher/student
  approach for surgical phase recognition.
\newblock {\em arXiv preprint arXiv:1812.00033}, 2018.

\bibitem{zappella2013surgical}
L.~Zappella, B.~B{\'e}jar, G.~Hager, and R.~Vidal.
\newblock Surgical gesture classification from video and kinematic data.
\newblock {\em Medical Image Analysis}, 17(7):732--745, 2013.

\bibitem{aaai2019ra}
H.~Zheng, L.~Yang, J.~Chen, J.~Han, Y.~Zhang, P.~Liang, Z.~Zhao, C.~Wang, and
  D.~Z. Chen.
\newblock Biomedical image segmentation via representative annotation.
\newblock In {\em AAAI}, 2019.

\bibitem{zhou2017fine}
Z.~Zhou, J.~Y. Shin, L.~Zhang, S.~R. Gurudu, M.~B. Gotway, and J.~Liang.
\newblock Fine-tuning convolutional neural networks for biomedical image
  analysis: Actively and incrementally.
\newblock In {\em IEEE Conference on Computer Vision and Pattern Recognition}.

\end{thebibliography}
\clearpage
\section{Appendix}
\vspace{-10mm}
\begin{table}[h]
	\centering
	\caption{Surgical workflow recognition performance of DBN and LRTD for active learning (mean$\pm$std., \%).
	}
	\resizebox{1\textwidth}{!}
	{
		\begin{tabular}{|l|l|c|c|c|c|c|c|}
			\hline
			\multicolumn{2}{|c|}{\begin{tabular}[c]{@{}c@{}}Methods\\\end{tabular}} & \begin{tabular}[c]{@{}c@{}}Data~ \\Amount\end{tabular} & \begin{tabular}[c]{@{}c@{}}Accuracy\\\end{tabular} & \begin{tabular}[c]{@{}c@{}}Precision\\\end{tabular} & \begin{tabular}[c]{@{}c@{}}Recall\\\end{tabular} & 
			\begin{tabular}[c]{@{}c@{}}Jaccard\\\end{tabular}  &
			\begin{tabular}[c]{@{}c@{}}F1 Score	\\\end{tabular}  \\
			\hline
			\multirow{9}{*}{Active Learning}
			& DBN~\cite{bodenstedt2019active}~
			& 10\%
			& ~$83.59\pm{8.30}$~
			& ~$\bm{80.61\pm{7.66}}$~
			& ~$82.71\pm{5.53}$~
			& ~$66.35\pm{9.43}$~
			& ~$79.25\pm{7.05}$~
			\\
			\cline{2-8}
			& LRTD (Ours)
			& 10\%
			& ~\bm{$83.98\pm{8.16}$}~
			& ~$80.10\pm{8.75}$~
			& ~\bm{$84.00\pm{3.43}$}~
			& ~\bm{$66.88\pm{9.34}$}~
			& ~\bm{$79.88\pm{6.91}$}~
			\\
			\cline{2-8}
			& DBN~\cite{bodenstedt2019active}
			& 20\%
			& ~$83.27\pm{8.43}$~
			& ~$80.73\pm{6.61}$~
			& ~$82.80\pm{6.53}$~
			& ~$66.42\pm{8.56}$~
			& ~$79.36\pm{6.18}$~
			\\
			\cline{2-8}
			& LRTD (Ours)
			& 20\%
			& ~$\bm{84.72\pm{7.90}}$~
			& ~$\bm{81.80\pm{5.99}}$~
			& ~$\bm{83.71\pm{5.20}}$~
			& ~$\bm{67.95\pm{9.16}}$~
			& ~$\bm{80.74\pm{6.65}}$~
			\\
			\cline{2-8}
			& DBN~\cite{bodenstedt2019active}
			& 30\%
			& ~$83.90\pm{9.05}$~
			& ~$81.41\pm{7.35}$~
			& ~$83.84\pm{6.07}$~
			& ~$67.81\pm{8.19}$~
			& ~$80.32\pm{5.85}$~
			\\
			\cline{2-8}
			& LRTD (Ours)
			& 30\%
			& ~$\bm{85.16\pm{7.88}}$~
			& ~$\bm{81.85\pm{6.88}}$~
			& ~$\bm{84.85\pm{4.39}}$~
			& ~$\bm{68.98\pm{8.60}}$~
			& ~$\bm{81.49\pm{6.03}}$~
			\\
			\cline{2-8}
			& DBN~\cite{bodenstedt2019active}
			& 40\%
			& ~$84.13\pm{8.56}$~
			& ~$\bm{82.38\pm{6.62}}$~
			& ~$84.00\pm{4.42}$~
			& ~$68.49\pm{7.72}$~
			& ~$80.92\pm{5.54}$~
			\\
			\cline{2-8}
			& LRTD (Ours)
			& 40\%
			& ~$\bm{85.00\pm{7.80}}$~
			& ~$81.11\pm{6.92}$~
			& ~$\bm{85.16\pm{3.70}}$~
			& ~$\bm{68.59\pm{8.54}}$~
			& ~$\bm{81.23\pm{6.01}}$~
			\\
			\cline{2-8}
			& DBN~\cite{bodenstedt2019active}
			& 50\%
			& ~$84.88\pm{8.35}$~
			& ~\bm{$83.00\pm{6.16}$}~
			& ~$83.54\pm{5.73}$~
			& ~$69.04\pm{7.59}$~
			& ~$81.34\pm{5.46}$~
			\\
			\cline{2-8}
			& LRTD (Ours)
			& 50\%
			& ~\bm{$85.87\pm{7.36}$}~
			& ~$82.76\pm{6.85}$~
			& ~\bm{$85.26\pm{4.30}$}~
			& ~\bm{$69.94\pm{8.99}$}~
			& ~\bm{$82.13\pm{6.22}$}~
			\\
			\hline
		\end{tabular}
	}
	\label{table:2}
	\vspace{-2mm}
\end{table}
\begin{table}[h!]
	\centering
	\caption{Comparison of our proposed LRTD method with state-of-the-art DBN~\cite{bodenstedt2019active} for active learning on various metrics of Jaccard, Precision and Recall, with results on each phase at different annotation ratios.}
	\resizebox{1\textwidth}{!}
	{
		\begin{tabular}{|c|c|c|c|c|c|c|c|c|}
			\hline
			\multirow{2}{*}{Method} & \multirow{2}{*}{Ratio}& \multirow{2}{*}{Preparation}  & CalotTriangle & Clipping    & Gallbladder & Gallbladder & Cleaning  & Gallbladder  \\
			& &  & Dissection & ~~~Cutting~~~ & Dissection &  Packaging & Coagulation & Retraction \\
			\hline
			\hline
			& & \multicolumn{7}{c|}{Jaccard} \\
			\hline
			DBN~\cite{bodenstedt2019active}    & 10\%  & 55.16 & $76.76$     & 73.52 & 77.25  & 64.01  & $\bm{59.00}$ & $\bm{58.75}$    \\
			\hline
			LRTD (Ours) & 10\%  & $\bm{57.67}$ & $\bm{77.84}$    & $\bm{74.24}$ & $\bm{77.41}$   & $\bm{64.04}$  & 58.29 & 58.65    \\
			\hline
			DBN~\cite{bodenstedt2019active}    & 20\%  & 56.18 & 76.15     & 72.43 & 76.41  & 64.25  & 59.67 & 59.84    \\
			\hline
			LRTD (Ours) & 20\%  & $\bm{57.52}$ & $\bm{77.92}$    & $\bm{75.24}$ & $\bm{78.26}$   & $\bm{66.77}$  & $\bm{59.98}$ & $\bm{60.93}$    \\
			\hline
			DBN~\cite{bodenstedt2019active}    & 30\%  & 58.81 & 77.19    & 73.38 & 77.47  & 66.26  & $\bm{60.59}$ & $\bm{60.94}$    \\
			\hline
			LRTD (Ours) & 30\%  & $\bm{61.98}$ & $\bm{79.19}$     & $\bm{75.14}$ & $\bm{78.66}$   & $\bm{67.22}$ & 60.10 & 60.56    \\
			\hline
			DBN~\cite{bodenstedt2019active}    & 40\%  & 59.34 & 77.03    & 74.10 & 77.43  & $\bm{67.71}$  & $\bm{61.89}$ & $\bm{61.97}$   \\
			\hline
			LRTD (Ours) & 40\%  & $\bm{61.68}$ & $\bm{78.89}$    & $\bm{74.98}$ & $\bm{78.41}$  & 64.31  & 60.01 & 61.88    \\
			\hline
			DBN~\cite{bodenstedt2019active}    & 50\%  & 60.95 & 78.33    & 72.94 & 78.30  & $\bm{67.71}$  & $\bm{62.42}$ & 62.66   \\
			\hline
			LRTD (Ours) & 50\%  & $\bm{61.28}$ & $\bm{80.45}$     & $\bm{78.12}$ & $\bm{79.26}$   & 66.74  & 60.70 & $\bm{63.03}$   \\
			\hline
			\hline
			& & \multicolumn{7}{c|}{F1 Score} \\
			\hline
			DBN~\cite{bodenstedt2019active}    & 10\%  & 70.10 & 86.65     & 84.87 & 86.53  & $\bm{79.32}$  & 72.95 & 74.30    \\
			\hline
			LRTD (Ours) & 10\%  & $\bm{72.00}$ & $\bm{87.29}$    & $\bm{86.20}$ & $\bm{87.02}$   & 79.01  & 72.95 & $\bm{74.72}$    \\
			\hline
			DBN~\cite{bodenstedt2019active}    & 20\%  & 71.10 & 86.14     & 83.84 & 85.88  & 79.44  & $\bm{74.11}$ & 75.01    \\
			\hline
			LRTD (Ours) & 20\%  & $\bm{72.15}$ & $\bm{87.40}$    & $\bm{86.28}$ & $\bm{87.46}$   & $\bm{81.63}$  & 73.82 & $\bm{76.45}$    \\
			\hline
			DBN~\cite{bodenstedt2019active}    & 30\%  & 72.98 & 86.84    & 84.08 & 86.63  & 80.95  & 74.82 & $\bm{75.93}$    \\
			\hline
			LRTD (Ours) & 30\%  & $\bm{75.55}$ & $\bm{88.15}$     & $\bm{86.23}$ & $\bm{87.86}$   & $\bm{81.78}$ & $\bm{74.96}$ & 75.92    \\
			\hline
			DBN~\cite{bodenstedt2019active}    & 40\%  & 73.24 & 86.74    & 84.96 & 86.49  & $\bm{82.13}$  & $\bm{75.99}$ & 76.89   \\
			\hline
			LRTD (Ours) & 40\%  & $\bm{75.31}$ & $\bm{87.92}$    & $\bm{86.31}$ & $\bm{87.67}$  & 79.46  & 74.54 & $\bm{77.44}$    \\
			\hline
			DBN~\cite{bodenstedt2019active}    & 50\%  & 74.10 & 87.59    & 84.39 & 87.22  & $\bm{82.17}$  & $\bm{76.55}$ & 77.36   \\
			\hline
			LRTD (Ours) & 50\%  & $\bm{75.23}$ & $\bm{89.01}$     & $\bm{88.16}$ & $\bm{88.05}$   & 81.10  & 75.26 & $\bm{78.10}$   \\
			\hline
			\hline
			& & \multicolumn{7}{c|}{Precision} \\
			\hline
			DBN~\cite{bodenstedt2019active}    & 10\%  & $\bm{77.14}$ & 87.82     &  87.07 & 88.26   & $\bm{81.06}$  & 69.68 & $\bm{73.16}$    \\
			\hline
			LRTD (Ours) & 10\%  & 71.17 & $\bm{89.33}$     & $\bm{87.35}$ &  $\bm{88.68}$   & 81.04  & $\bm{70.56}$ & 72.57    \\
			\hline
			DBN~\cite{bodenstedt2019active}    & 20\%  & $\bm{81.02}$ & $\bm{89.33}$     &  79.87 & 87.14   & 73.05  & 75.05 & 73.15    \\
			\hline
			LRTD (Ours) & 20\%  & 75.39 & 87.48     & $\bm{86.01}$ &  $\bm{89.06}$   & $\bm{81.96}$  & $\bm{75.73}$ & $\bm{76.96}$    \\
			\hline
			DBN~\cite{bodenstedt2019active}    & 30\%  & 78.36 & $\bm{89.56}$     &  84.33 & 85.67   & 76.30  & $\bm{73.31}$ & $\bm{77.58}$    \\
			\hline
			LRTD (Ours) & 30\%  & $\bm{78.40}$ & 88.64     & $\bm{85.62}$ &  $\bm{89.34}$   & $\bm{84.18}$  & 72.75 & 73.99    \\
			\hline
			DBN~\cite{bodenstedt2019active}    & 40\%  & $\bm{77.36}$ & $\bm{89.65}$     &  $\bm{90.18}$ & 85.52   & $\bm{82.35}$  & $\bm{72.53}$ & $\bm{79.09}$    \\
			\hline
			LRTD (Ours) & 40\%  & 76.57 & 89.49     & 85.27 &  $\bm{88.85}$   & 76.65  & 72.52 & 78.45   \\
			\hline
			DBN~\cite{bodenstedt2019active}    & 50\%  & $\bm{82.14}$ & 88.66     &  88.21 & 86.79   & $\bm{84.26}$  & 73.05 & 77.86    \\
			\hline
			LRTD (Ours) & 50\%  & 76.92 & $\bm{90.12}$     & $\bm{90.52}$ &  $\bm{88.60}$   & 78.50  & $\bm{74.44}$ & $\bm{80.23}$    \\
			\hline
			\hline
			& & \multicolumn{7}{c|}{Recall} \\
			\hline
			DBN~\cite{bodenstedt2019active}    & 10\%  & 72.57 & $\bm{86.91}$     & 85.41 & 87.53   & $\bm{80.15}$  & $\bm{83.59}$ &  82.80   \\
			\hline
			LRTD (Ours)& 10\%  & $\bm{80.45}$ & 86.89     &  $\bm{86.43}$ & $\bm{87.72}$   & 79.83  & 82.67 &  $\bm{84.00}$    \\
			\hline
			DBN~\cite{bodenstedt2019active}    & 20\%  & $\bm{76.59}$ & 83.51     & 84.06 & $\bm{90.69}$   & $\bm{85.06}$  & $\bm{82.96}$ &  75.63    \\
			\hline
			LRTD (Ours) & 20\%  & 76.45 & $\bm{88.64}$     &  $\bm{88.52}$ & 88.34   & 84.13  & 78.10 &  $\bm{81.85}$    \\
			\hline
			DBN~\cite{bodenstedt2019active}    & 30\%  & 71.10 & 86.14     & 85.50 & $\bm{89.10}$   & $\bm{83.76}$  & $\bm{86.12}$ &  $\bm{85.20}$    \\
			\hline
			LRTD (Ours) & 30\%  & $\bm{77.95}$ & $\bm{89.09}$     &  $\bm{88.79}$ & 88.76   & 82.01  & 82.43 &  84.94    \\
			\hline
			DBN~\cite{bodenstedt2019active}    & 40\%  & 76.71 & 85.55     & 83.04 & $\bm{90.69}$   & 83.62  & $\bm{86.04}$ &  $\bm{82.34}$    \\
			\hline
			LRTD (Ours) & 40\%  & $\bm{80.17}$ & $\bm{87.71}$     &  $\bm{88.83}$ & 88.67   & $\bm{86.24}$  & 82.34 &  82.17    \\
			\hline
			DBN~\cite{bodenstedt2019active}    & 50\%  & $\bm{72.90}$ & 87.73     & 82.96 & $\bm{90.44}$   &81.46  & $\bm{85.62}$ & $\bm{83.65}$    \\
			\hline
			LRTD (Ours) & 50\%  & 79.18 & $\bm{88.95}$     &  $\bm{87.23}$ & 89.83   & $\bm{87.91}$  & 81.35 &  82.31    \\
			\hline
		\end{tabular}
	}
	\label{phaseresults}
	\vspace{-3mm}
\end{table}
\begin{figure}[ht]
	\centering
	\includegraphics[width=0.8\linewidth]{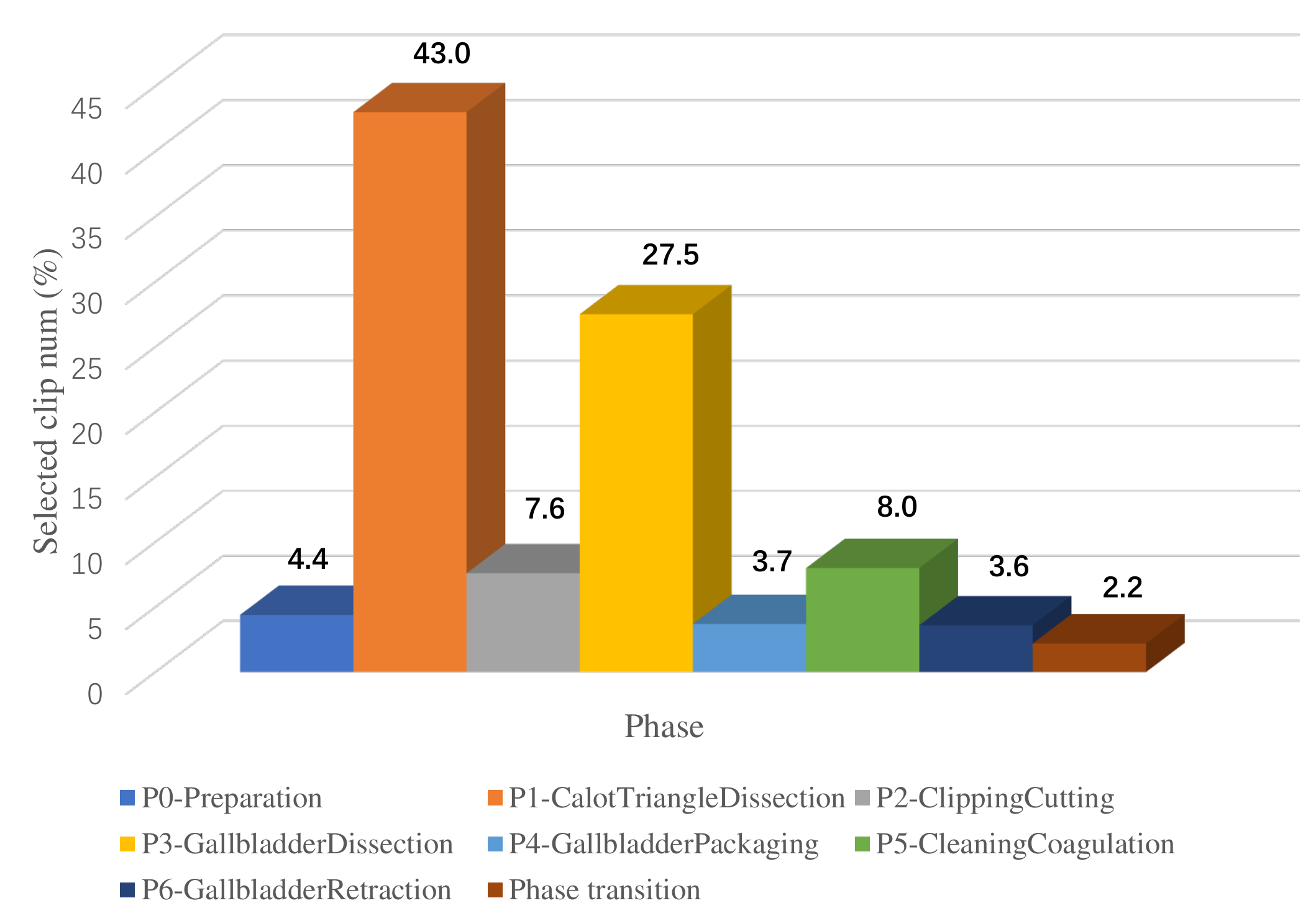}
	\caption{Ratio statistics about selected clips' phases.}
	\label{fig:histogram}
\end{figure}

\end{document}